\documentclass[letterpaper, 10 pt, conference]{ieeeconf}  

\IEEEoverridecommandlockouts                              

\usepackage{cite}
\usepackage{hyperref}
\usepackage{multirow}
\usepackage{graphicx}
\usepackage{gensymb}
\usepackage{acronym}
\usepackage{siunitx}
\usepackage{amsmath}
\usepackage{amssymb}
\usepackage{tikz}
\usepackage{xcolor}
\usepackage{amsfonts}
\usepackage{microtype}

\acrodef{RL}{Reinforcement Learning}
\acrodef{NN}{Neural Network}
\acrodef{DoF}{Degree of Freedom}
\acrodef{ID}{Inverse Dynamics}
\acrodef{IMU}{Inertial Measurement Unit}
\acrodef{PPO}{Proximal Policy Optimization}
\acrodef{PSD}{Power Spectral Density}
\acrodef{MLP}{Multi-Layer Perceptron}
\acrodef{COG}{Center Of Gravity}
\acrodef{EE}{end effector}
\acrodef{AoA}{Angle of Arrival}
\acrodef{RF}{radio frequency}
\acrodef{SDR}{Software Defined Radio}
\acrodef{SM}{Switching Matrix}
\acrodef{CW}{Continuous Wave}
\acrodef{ISM}{Industrial, Scientific, and Medical}
\acrodef{TDM}{Time Division Multiplexing}

\DeclareSIUnit{\rad}{rad}

\title{\LARGE \bf
Obstacle-Avoidant Leader Following with a Quadruped Robot
}

\author{Carmen Scheidemann$^1$, Lennart Werner$^1$, Victor Reijgwart$^2$, Andrei Cramariuc$^1$,\\Joris Chomarat$^1$, Jia-Ruei Chiu$^1$, Roland Siegwart$^2$, Marco Hutter$^1$
\thanks{Authors are members of the $^1$Robotic Systems Lab and $^2$Autonomous Systems Lab, ETH Z{\"u}rich, Switzerland. Email: {\tt\small carmensc@ethz.ch}}%
\thanks{This project has received funding from armasuisse Science and Technology and the European Union’s Horizon Europe Framework Programme under grant agreement No 101121321. This work has been conducted as part of ANYmal Research, a community to advance legged robotics.}%
}

\begin{document}

\maketitle
\thispagestyle{empty}
\pagestyle{empty}

\begin{abstract}
Personal mobile robotic assistants are expected to find wide applications in industry and healthcare. For example, people with limited mobility can benefit from robots helping with daily tasks, or construction workers can have robots perform precision monitoring tasks on-site. However, manually steering a robot while in motion requires significant concentration from the operator, especially in tight or crowded spaces. This reduces walking speed, and the constant need for vigilance increases fatigue and, thus, the risk of accidents. This work presents a virtual leash with which a robot can naturally follow an operator. We use a sensor fusion based on a custom-built RF transponder, RGB cameras, and a LiDAR. In addition, we customize a local avoidance planner for legged platforms, which enables us to navigate dynamic and narrow environments. We successfully validate on the ANYmal platform~\cite{Hutter2016ANYmalRobot} the robustness and performance of our entire pipeline in real-world experiments. 
The video is available at: \href{https://leggedrobotics.github.io/obstacle-avoidant-leader-following/}{obstacle-avoidant-leader-following}. 

\end{abstract}

\section{Introduction}

With mobile robots becoming more ubiquitous in industrial and personal applications, autonomous human following is a significant step towards increased operator freedom. Such capabilities already exist in many consumer-grade unmanned aerial vehicles~\cite{Mondragon20113DVehicles, Chen2019TheUAV, DJI_2024}, e.g., following and filming the user during an activity, such as hiking or riding a bike. Similar human-following capabilities would benefit the operation of ground-based robots in various scenarios. Speed and efficiency would increase in challenging environments (e.g., crowds or narrow spaces), as the operator would not have to alternate between moving the robot and themselves. Additionally, the operator would require less concentration and skill, making the robot accessible to a broader range of, potentially untrained, users. In particular, it would enable people with physical disabilities (i.e., wheelchair users) to operate and have personal assistance robots without requiring a complex physical controller~\cite{scheidemann2024cybathlonleggedmobile}.

Most existing human-following solutions for mobile ground robots either rely on tethered connections between the robot and user (\textit{i.e.}, a physical leash)~\cite{Young2011HowInterface}, use sensors to track a transmitter on the person~\cite{Fang2010ABeacon, Afghani2013FollowBeacons, li2022quadruped}, use purely image-based tracking~\cite{rollo2024continuous}, or follow a human by using a range measurement sensor and maintaining a constant distance to the nearest obstacle in front of them, predicated on the assumption that this is their  human~\cite{Tripathi2021HumanSensor, Yang2019ASystem}.

\begin{figure}[!t]
    \centering
    \includegraphics[width=0.95\linewidth]{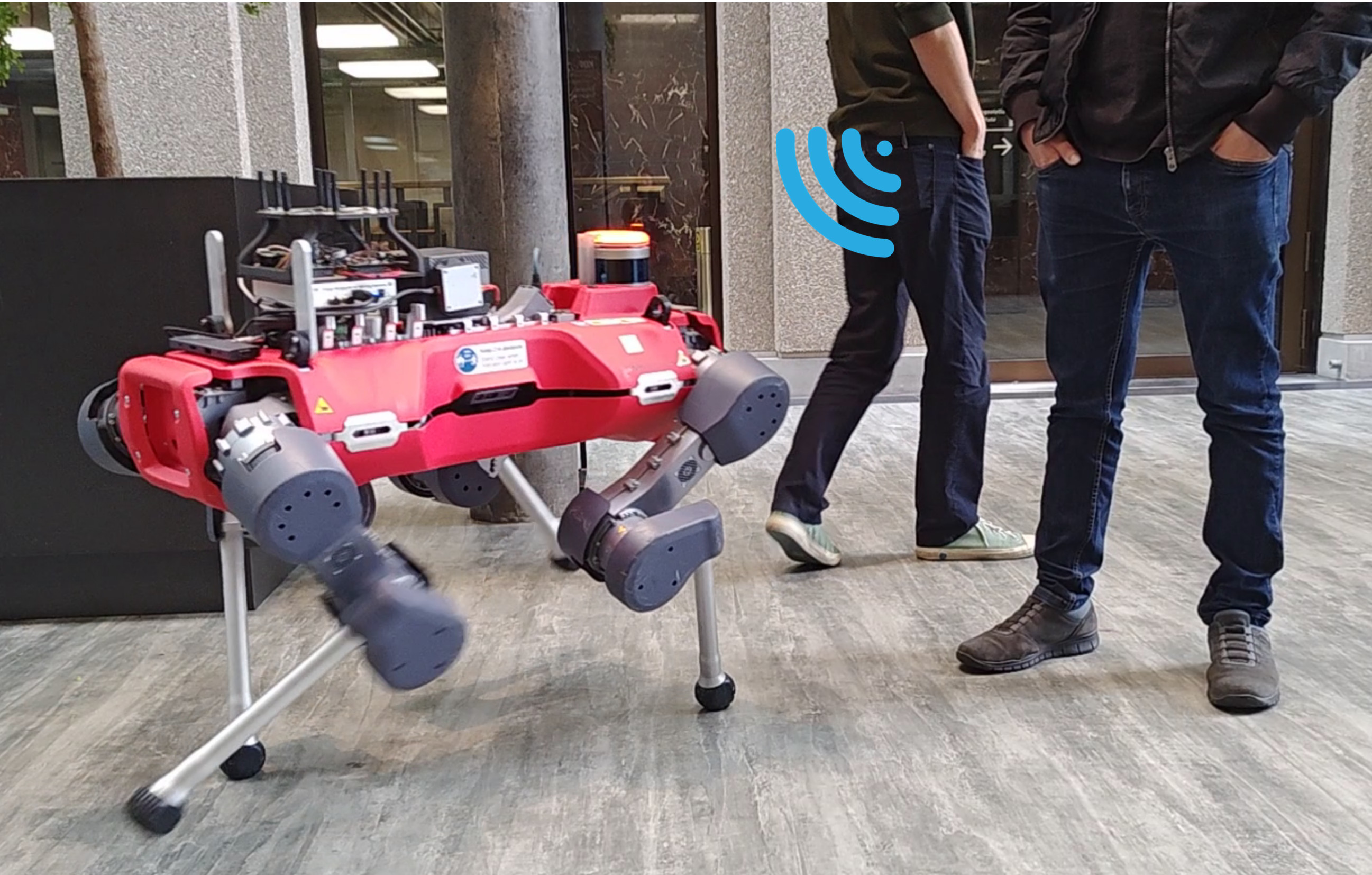}
    \caption{We deploy our full pipeline on the ANYmal robot and show resilient leader following in crowded and complex environments. The robot identifies the leader through a beacon they carry, emphasized with blue, the signal of which is picked up by a custom antenna array on the robot.}
    \label{fig:teaser}
    \vspace{-0.5cm}
\end{figure}

These systems do not incorporate appropriate path planning and obstacle avoidance systems that could deal with complex environments, including crowds of moving people or narrow gaps such as doors. While the human operator implicitly does the high-level planning and, in many cases, clears a path through crowds, fast reactive behavior and smart local planning are needed to maintain efficiency and safety. Irrespective of path planning algorithms, most algorithms do not actively track the leader, but only a fixed distance or a beacon. Due to multipathing and signal interference, wireless beacons on their own are often not robust enough without the assistance of other sensors.

\begin{figure*}[!t]
\centering
\includegraphics[width=1\linewidth]{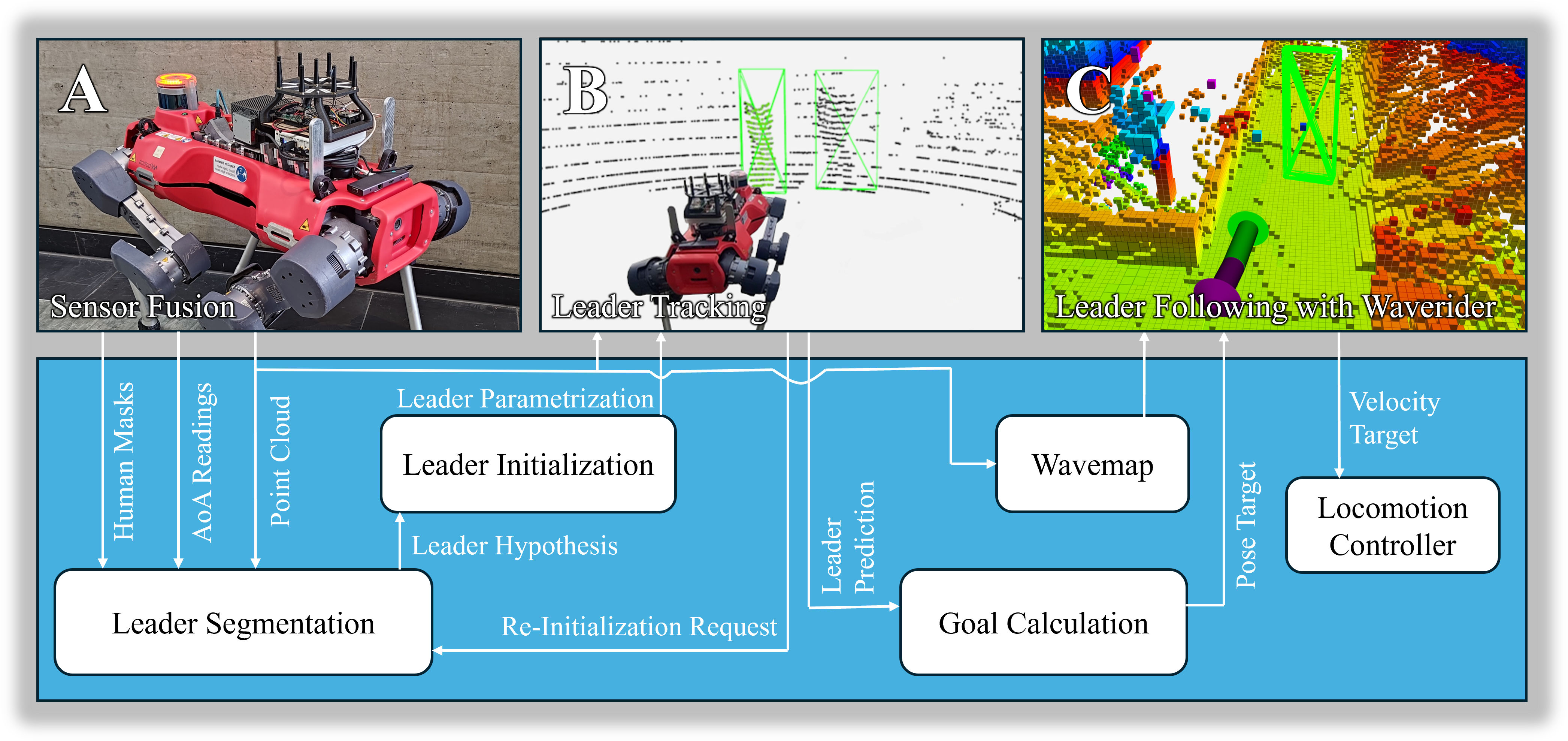}
\vspace{-0.6cm}
\caption{The presented pipeline consists of three main elements: the sensor fusion (A), leader tracking (B), and leader following (C). We combine measurements from the onboard cameras and LiDAR unit of the robot with an additional custom \acf{AoA} sensor unit. From these measurements, we segment our leader out of the scene, which may involve other people. Using an EKF, we track the leader's motion, which allows us to keep track of them, even when occluded. By adding a new waverider policy for dynamic obstacles, the robot can follow the leader through crowded spaces, avoiding collision with both the environment and (potentially moving) other people.}
\label{fig:pipeline_overview}
\vspace{-0.3cm}
\end{figure*}

This work aims to enable dynamic person following using ANYmal~\cite{Hutter2016ANYmalRobot}, a quadrupedal robotic platform, shown in Figure~\ref{fig:teaser}. We present a full pipeline from human recognition to tracking and following, allowing the user to decide at what distance and orientation the robot should follow them. Our approach is a novel multimodal fusion that solves multiple challenges in the current state of the art. First, we combine cameras and depth sensors to detect people and track the leader. To disambiguate in crowded situations or after briefly losing line of sight, we additionally develop and integrate a custom Angle of Arrival (AoA) sensor so that the leader can carry a beacon that will identify them from others. Finally, we also tackle the local obstacle avoidance challenge by adapting an existing method, waverider~\cite{Reijgwart2024WaveriderAvoidance}, to ground-based robots. We also enable waverider to actively avoid dynamic obstacles by explicitly integrating them as seperate Riemannian Motion Policies (RMPs). This allows our system to navigate challenging environments and adapt its relative position to the leader depending on need. Ultimately, operator efficiency is increased, and we present a full framework for autonomous robot control through leader following. To summarize, our contributions are as follows:
\begin{itemize}
    \item A multi-modal sensor fusion pipeline for leader tracking that can deal with temporary occlusions and crowds.
    \item A novel Angle of Arrival (AoA) sensor for robots to accurately track a beacon, even in complex environments.
    \item Integration of a reactive local obstacle avoidance planner for ground-based robots to reactively.
\end{itemize}

\section{Related Work}
\label{chp:related}

Fang \textit{et al.}~\cite{Fang2010ABeacon} employed two radio frequency (RF) antennas on their robot to triangulate the orientation of an RF transmitter carried by the target person. The distance to the target was measured using a sonar system with eight ultrasound sensors arranged in a semicircle. Their approach does not address temporary occlusions because proximity is assumed, nor is it robust to complex environments, as the RF sensor suffers from multipath interference. A similar approach was pursued by Afghani \textit{et al.}~\cite{Afghani2013FollowBeacons}, who used a combination of infrared (IR) receivers on the robot and a transmitter carried by the person. Their approach has similar deficiencies, with no sensor redundancy in tracking, thus only being functional in rudimentary environments with no obstacle avoidance.

Few previous works exist on ground robots that actively track and follow a human could be found. Pe\~{n}a \textit{et al.}~\cite{Pena2019FollowOptimization} developed a human-following robot using a camera to track a triangle formation of three IR transmitters on the target's back instead of the human directly. This approach is intrusive and perspective-limited (tracking is only possible from behind). Li~\textit{et al.}~\cite{li2022quadruped} use a quadrupedal robot to track a beacon, using externally mounted base stations on-site. 
Their approach limits the deployment range to the area spanned by the beacons, severely limiting deployability and scalability. Having all required sensors mounted directly onto the robot allows us to overcome such limitations directly. A purely image-based leader tracking was proposed by Rollo \textit{et al.}~\cite{rollo2024continuous}, however, it was not shown to work in dynamic or crowded environments.
Several image-based leader tracking approaches exist for aerial vehicles where issues on perspective, line of sight, and path planning are much simplified~\cite{Mondragon20113DVehicles, Chen2019TheUAV}. Commercial approaches to leader tracking on a legged robot~\cite{Unitree_2024} and aerial vehicles~\cite{DJI_2024} exist, showcasing their relevance. However, no information exists on the underlying implementation. 

Other work, such as Tripathi \textit{et al.} and Yang \textit{et al.}~\cite{Tripathi2021HumanSensor, Yang2019ASystem}, carried no transmitter. Instead, sonar systems were used to identify the target's legs. However, these systems are severely limited to flat environments without crowds. Young \textit{et al.} proposed a tethered approach akin to a dog leash to improve the intuitiveness of human-robot interaction~\cite{Young2011HowInterface}. A retractable leash was used with a rotary encoder to track its orientation, allowing the robot to follow directly behind, at an angle, or in front of the leader. Due to the robot's lack of obstacle avoidance capabilities, a high burden was placed on the user in challenging environments. In contrast to previous approaches, we entirely automate the leader following, providing planning, obstacle detection, and a novel beacon for re-detection in case of obstruction or other issues.

\section{Overview}

Our pipeline consists of three main elements, which can be seen in Figure~\ref{fig:pipeline_overview}: sensor fusion (A), leader tracking (B), and leader following (C). To add robustness and redundancy, we combine measurements from the onboard LiDAR (Velodyne VLP-16), a custom \acf{AoA} sensor array, and four Intel Realsense 265 RGB-D cameras for sensor fusion. The leader carries a pocket-sized RF transmitter that our \ac{AoA} sensor can track, thus constantly providing the robot with a leader heading even when other modalities fail. Using the LiDAR and camera, we track the leader with an Extended Kalman Filter (EKF) that uses a constant velocity model to forward predict the leader's position. Using a quadrupedal robot, ANYmal, we follow the leader at a set distance and angle based on the EKF prediction, diverging from it when necessary to avoid static and dynamic obstacles.

\subsection{Sensor Fusion}
\label{sec:sensor-fusion}

We fuse the readings from all three types of sensors to obtain our initial leader hypothesis. We run YOLOv8~\cite{redmon2016lookonceunifiedrealtime} to segment out every human from all four camera feeds individually. While multiple methods for human detection exist, using either LiDARs~\cite{milioto2019rangenet++} or cameras~\cite{redmon2016lookonceunifiedrealtime, he2017mask}, most are too computationally expensive to run real-time on mobile platforms. Additionally, the low perspective of the LiDAR on ANYmal produced poor-quality results with state-of-the-art segmentation networks due to said detectors being mainly trained on autonomous driving datasets with higher and further away viewpoints. Our choice of YOLOv8 as a detection network was motivated by its higher semantic content of images and low runtime, allowing us to run it in real-time on the onboard Jetson Orin.

We compare the predicted human bounding boxes to the \ac{AoA} reading to find the person that corresponds most closely to the angle reading. This process is repeated until a human bounding box is found within $20^{\circ}$ degrees of the leader angle reported by the \ac{AoA} sensor. Once a valid leader is identified, we reproject the bounding box into the LiDAR point cloud to extract a hypothesis of the leader position relative to the robot. To initialize the full 3D position, we assume the human is in the foreground and extract the first large mass the box intersects.

\subsection{Leader Tracking}
\label{sec:leader-tracking}


Human tracking is extensively studied using both cameras~\cite{Wojke2018deep}, LiDAR~\cite{qi2020p2b, yin2021center}, or a fusion of both modalities~\cite{kim2021eagermot, meng2023hydro, wang2022deepfusionmot, zhao2020dynamic}. However, many of these methods are computationally expensive as they fuse information in large networks or require object detection to run separately in both sensor modalities. Methods developed for mobile robotics applications instead often utilize cheaper detectors with a combination of filtering and clustering-based approaches~\cite{Liu2016PeopleRobots, Wagh2014HumanSystem}. Our use of a lightweight beacon carried by the leader adds a layer of robustness to occlusions, allowing us to use more computationally inexpensive methods.

Inspired by literature, we track our leader using an Extended Kalman Filter (EKF), which uses a linear motion model. 
To initialize the EKF, we wait for two consistent leader hypotheses to filter out outliers. From these two initial matches, we derive an initial guess on the leader velocity in the $xy$ plane of the robot. The resulting leader parametrization, consisting of a position $p_0^l\in\mathbb{R}^2$, velocity $v_0^l\in\mathbb{R}^2$, and a set of 3D LiDAR points on the leader $P_0^l\in\mathbb{R}^{n\times3}$, are used to initialize leader tracking.
We use the prediction step from the EKF together with the motion estimation of the robot to project the previously detected leader points $P^l_{t-1}$ into the current time frame to obtain the predicted leader point cloud $\hat{P}^l_t$. Before matching them with the latest LiDAR scan $P_t$, we remove large planes (i.e., walls and the floor) using RANSAC~\cite{fischler1981random} to eliminate 3D structures we are certain are not humans. We then select new leader points through a KD-tree nearest-neighbor search~\cite{bentley1975multidimensional} by matching each point in $\hat{P}^l_t$ with the nearest points in $P_t$. After each search association step, we reject 40\% of the points furthest from the predicted leader center to obtain the final measured $P^l_t$. This heuristic prevents the leader point cloud from spreading onto adjacent static objects or people while not degrading the performance of the EKF. When available, we fuse the image segmentation to bound the association step. When the leader is occluded or out of view of the cameras, we rely solely on the EKF motion model.

While the \ac{AoA} sensor can validate our target, it can not provide a 3D position for the robot to track, which motivates our need for an EKF. However, the \ac{AoA} and EKF prediction angles are constantly monitored. Leader selection is re-initialized if the two diverge for longer than a predefined amount of time. If this happens, the robot moves towards its last known target, where it interrupts movement until a new leader candidate is found through the process described in Section~\ref{sec:sensor-fusion}.

\subsection{Leader Following}
\label{sec:leader-following}

We aim to have the robot follow the leader at a set distance and angle with respect to the leader's heading. Accordingly, we calculate the robot's position goal using the leader's pose and velocity to determine a fixed set point next to the leader at which the robot should aim to be. Assuming the pilot is in motion, this target is constantly moving. Rather than trying to exactly mirror the pilot motions, which might cause collisions (e.g., dynamic obstacles or walking next to a wall), we adapt and use a local avoidance planner (see Section~\ref{sec:waverider}).

\begin{figure}
    \centering
    \includegraphics[width=0.95\linewidth]{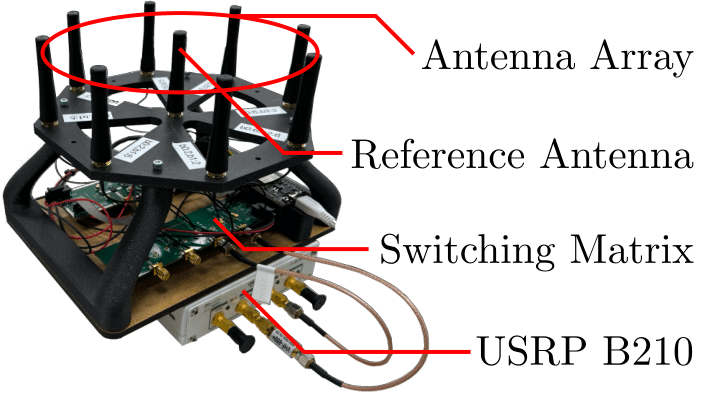}
    \caption{Our novel \acf{AoA} Sensor composed of a \acf{SDR}, \acf{SM} and an antenna array on top.}
    \label{fig:aoa-array-closeup}
\end{figure}

\section{\ac{AoA} Sensor}
\label{sec:aoa-sensor}
Due to the lack of affordable, commercially available \SI{360}{\degree} \ac{AoA} sensors, a custom sensor was implemented for this project. The system functions by locating the wave incidence angle of a \ac{CW} beacon within the \SI{2.4}{\GHz} \ac{ISM} band. A small low-power ($\leq$ \SI{0}{\dB \m}) handheld signal generator is used as a target beacon for our experiments. Our system can predict the \ac{AoA} with sufficient precision under heavy indoor multi-pathing conditions using an affordable two-channel coherent \ac{SDR} with \ac{TDM}. It is described in full detail in concurrent work \cite{werner2025directionfindingsoftwaredefined}.

\subsection{Hardware}
Figure~\ref{fig:aoa-array-closeup} highlights the components and placement used for the sensor, and in Figure~\ref{fig:pipeline_overview}A the sensor is mounted on the robot. An 8+1 Uniform Circular Array with an antenna spacing of approximately half a wavelength $\lambda/2$ is mounted in direct line of sight to the beacon. The eight outer antennas are multiplexed through a \ac{SM} and connected to a receiving port of the \ac{SDR} (Ettus USRP B210 \cite{ettusUSRPB210}). As a permanent reference for phase synchronization of the \ac{TDM} array, the central antenna remains permanently connected to the other coherently sampled receive channel of the \ac{SDR}.

\begin{figure*}[!t]
\centering
\includegraphics[width=0.95\linewidth]{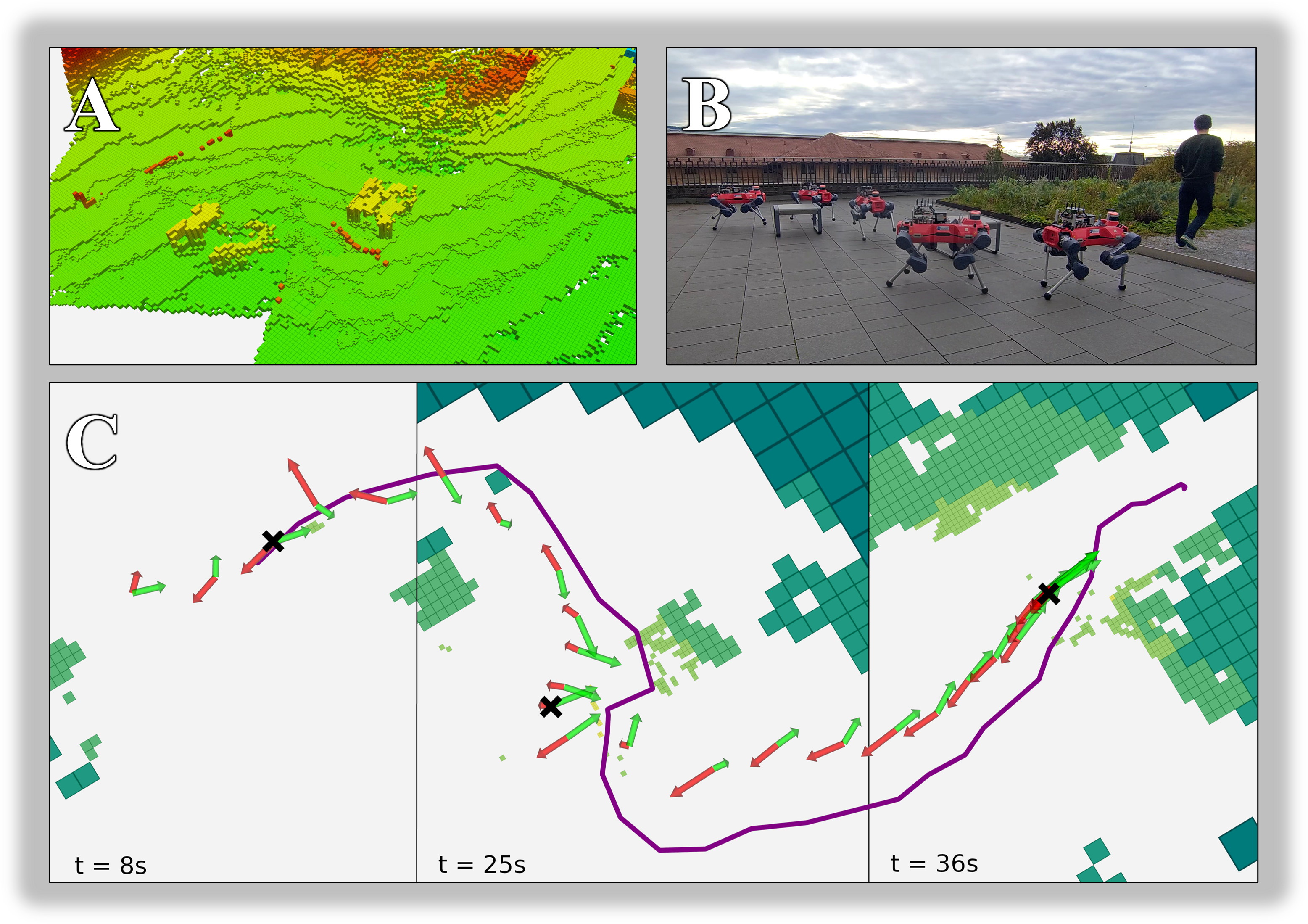}
\vspace{-10pt}
\caption{We modify waverider to track our leader while autonomously avoiding static obstacles, such as benches and door frames. We create an occupancy map with wavemap, actively filtering out human points to only include the static scene. The map resulting from a tracking sequence can be seen in (A). Said sequence is visualized in (B), with only the final leader frame visualized to reduce visual clutter. In (C), we show the path the robot takes while avoiding collisions. The leader trajectory (as recorded by the robot) is visualized via the purple line and the robot trajectory via the sequence of arrow clusters. Each cluster represents a pose and the forces acting on the robot during that single time step, the green symbolizing the attractive force of the goal and the red the repelling forces of the obstacles within range. The image is split into three frames to highlight the multi-resolution nature of waverider, with the green blocks representing the obstacles. Each third of the image represents one timestep, and the black cross shows the exact robot position for that timestep.}
\label{fig:wavemap_big}
\vspace{-10pt}
\end{figure*}

\subsection{Signal Processing}
To avoid the high costs associated with an eight channel coherent \ac{SDR} a \ac{TDM} sampling of the antennas is used. As such, in each measurement cycle, the antennas are sampled sequentially. Since the \ac{CW} beacon transmits a cyclostationary signal, a pseudo-coherent measurement can be reconstructed through time synchronization. Jitter in the switching and sampling with resulting phase shifts significantly deteriorates all \ac{AoA} estimation algorithms.
To minimize these effects, we use a second antenna, coherently sampled, as a reference for synchronization. The pseudo coherent signal $X_{\text{s, i}}$ is calculated from the array antenna signal $X_i$ and the reference antenna $X_r$ as shown in Equation~\eqref{eq:phaseAlign}.
$X_r^*$ is the complex conjugate of the reference signal, with $\cdot$ being the element-wise multiplication operator.
\begin{align}
    \mathbf{X_{\text{s, i}}} &= \mathbf{X_i} \cdot \mathbf{X_r^*}
    \label{eq:phaseAlign}
\end{align}

Due to the superior properties regarding stability, accuracy, and the possibility of coherent signal separation~\cite{lavate2010performance}, the MUSIC algorithm~\cite{schmidt1986multiple} is used for \ac{AoA} estimation.
Our approach extends the Dual Channel Cyclic Music algorithm~\cite{8695705} with spatial smoothing~\cite{17496} to decorrelate 
coherent signals present in strong multi-path environments.
Spatial smoothing for Uniform Circular arrays additionally requires the transformation of the signal vector into a Vandermonde structure, which is amenable for spatial smoothing~\cite{299559}.
With eight antennas, we create three sub-arrays for one expected signal to get the highest immunity to multi-pathing from two simultaneous directions. We also treat an arbitrary amount of uncorrelated signals as noise.

With the presented structure, we achieve high precision predictions in high noise and multipathing indoor scenarios at an update rate of \SI{5}{\Hz}. We also add a three-step median filter on the measurement to remove noise that can happen if there are too many multipath effects in the environment. The resulting frequency is too low to solely provide the robot with a target. However, the EKF described in Section~\ref{sec:leader-tracking} runs at a frequency of \SI{10}{\Hz}, providing smoother targets.


\section{Local Navigation}
\label{sec:waverider}
Classical state-of-the-art approaches to local obstacle avoidance combine a 3D representation of the world with either search~\cite{mohta2018fast}, sampling~\cite{karaman2011sampling}, or optimization-based~\cite{oleynikova2020open} methods. However, these methods suffer from performance bottlenecks in the mapping or planning loop, sometimes requiring prior maps of the environment, making them unfeasible in live scenarios. 
In this work, we extend waverider~\cite{Reijgwart2024WaveriderAvoidance}, a reactive avoidance planner for UAVs, which uses a hierarchical map structure to formulate Riemannian Motion Policies (RMPs)~\cite{Ratliff2018RiemannianPolicies}. We adapt waverider to work on ground robots and avoid dynamic obstacles. Unlike other works, this approach is computationally efficient ($2.5$ CPU cores) and runs at a high frequency (\SI{50}{\Hz}).

The navigation system is formulated as a collection of objectives, represented and combined using the mathematical framework of RMPs. We use the following policies: 1) a goal-seeking policy that pulls the robot towards a waypoint alongside its leader; 2) a yaw policy that encourages the robot to face toward the goal; 3) a static obstacle avoidance policy to avoid collisions with the environment; and 4) a dynamic obstacle avoidance policy that avoids collisions with humans, to safely navigate crowds. Each policy generates a desired acceleration vector and a corresponding Riemannian metric that quantifies its directional importance. The policies are combined into a single reference acceleration in $SE(2)$ by computing their metric-weighted average, which is then integrated and sent to the locomotion controller as a reference velocity. A key advantage of RMPs is their modularity, allowing the presented navigation system to be extended to other tasks by simply formulating and including additional objectives as RMPs.

The goal and yaw policies are formulated as attractor policies~\cite{Ratliff2018RiemannianPolicies}, whose targets are the goal's position in $\mathbb{R}^2$ and the yaw angle of the bearing vector from the robot to the goal in $SO(2)$, respectively. In addition to proportional terms that pull the robot to walk toward and face the goal, each policy includes a derivative term that smooths out the motion.

To avoid collisions, we use wavemap~\cite{Reijgwart2023EfficientCompression} to efficiently fuse all depth images from the robot's front, left, right, and back-facing depth cameras into a single map. An obstacle avoidance policy is then derived from this map using waverider~\cite{Reijgwart2024WaveriderAvoidance}. Waverider leverages the hierarchical structure of wavemap's occupancy maps to summarize the obstacles surrounding the robot as multi-resolution cubes. This allows the policy to safely avoid very small obstacles near the robot while also efficiently considering distant obstacles at a lower resolution, which helps it avoid getting stuck in local minima. 

3D occupancy mapping methods, including wavemap, assume that the environment is static to avoid excessive computational costs. Although they can be tuned to better show dynamic obstacles, reducing the delay for objects to appear and fade from the map typically comes at the cost of increasing the map's overall noise levels. Therefore, we propose to avoid this inherent trade-off by explicitly detecting objects that are expected to move and handle them separately. In particular, we reuse the detection from YOLOv8~\cite{redmon2016lookonceunifiedrealtime} from Section~\ref{sec:sensor-fusion}. The detections are then converted to 3D Axis Aligned Bounding Boxes and removed from the depth images so that wavemap only reconstructs the remaining static environment. The bounding boxes are, in turn, accumulated into a rolling buffer and processed by the dedicated, dynamic obstacle avoidance policy. Aside from separating the concerns of delays and map quality, this also allows the dynamic obstacle avoidance policy to be tuned separately. For example, to keep a larger safety distance from people than objects.

Although the static and dynamic obstacle avoidance policies are formulated in $\mathbb{R}^3$, we convert them into $SE(2)$ using the RMPs pullback operator~\cite{Ratliff2018RiemannianPolicies} before combining them with the goal and yaw policies. For this work, we do not control the robot's roll, pitch, and height. However, pulling the policies into $SE(3)$ instead would allow the robot to traverse over or under obstacles autonomously.

\section{Results}

We deploy our pipeline on the ANYmal robot and perform leader-following tests in multiple scenarios to validate our approach. Additionally, we provide quantitative or qualitative evaluations for the individual components of our pipeline.

\subsection{AoA Sensor}

We evaluate the performance of the AoA sensor in different use cases, showing its usefulness both indoors and outdoors. For a quantitative analysis we measure the AoA readings for a fixed location of the beacon over time periods of one minute. First, we evaluate the optimal scenario: an open space and the beacon in the same $x-y$ plane as the antenna array. We obtain a mean error of $3^{\circ}$ and a standard deviation of $11^{\circ}$ for various angles. All orientations yield approximately the same error, as is expected, due to the symmetric nature of our beacon in the $x-y$ plane. We observe a slightly decreased performance for the same angles if the beacon is moved 40cm upwards out of a plane. This is to be expected due to the geometry of the antenna array and the assumptions we make for the signal processing described in Section~\ref{sec:aoa-sensor}. In this scenario, we measure a mean error of $4^{\circ}$ with a standard deviation of $14^{\circ}$. As our sensor can handle multi-path effects, we also test in the more challenging scenario of the robot standing directly next to a wall. Here, the mean error grows to $7^{\circ}$ with a standard deviation of $27^{\circ}$. With minimal filtering, this level of accuracy still proves sufficient to differentiate between multiple people who are in proximity to each other.

\subsection{Local Navigation}
We evaluate our extensions to waverider~\cite{Reijgwart2024WaveriderAvoidance} which allow it to operate on ground-based robots and dynamic scenes. We show qualitatively in multiple scenarios that the robot is able to achieve its goal while avoiding both static and dynamic obstacles.\footnote{The accompanying \href{https://leggedrobotics.github.io/obstacle-avoidant-leader-following/}{video} showcases the leader selection and dynamic local navigation, as well as presenting additional details on the experiments.} It also deals with adversarial behavior from humans trying to block its way. The full navigation pipeline runs on only $2.5$ cores of the onboard computer while replanning at a constant rate of \SI{50}{\Hz}. Visualizations of the resulting 3D reconstruction of wavemap~\cite{Reijgwart2023EfficientCompression}, from which waverider generates its policies, are shown in Figure~\ref{fig:wavemap_big}.

\subsection{Leader Following}
We show the robustness of our leader-following pipeline in multiple scenarios designed to emulate common challenges that other systems do not tackle:

\subsubsection{Environment with Obstacles} First, we show our system in a static outdoor environment with two placed obstacles, between which the leader walks (the scene is shown in Figure~\ref{fig:wavemap_big}B). The robot successfully follows the leader while dodging the obstacles. The resulting hierarchical 3D reconstruction from Wavemap is shown in Fig~\ref{fig:wavemap_big}A, where only the highest resolution is displayed. Important to note is that even the thin legs of the chairs used as obstacles are visible in the map, as is the nearby vegetation. Only a few spurious points from the human are visible, showing that even computationally inexpensive methods are sufficient to obtain clean maps. The resulting map is used by waverider, as visualized in Figure~\ref{fig:wavemap_big}C, to plan the robot trajectory. The figure superimposes some of the obstacle hierarchies used by waverider at different timesteps into a single image. It shows the resulting forces applied to the robot from the different RMP policies, in the form of attractor/repulsor arrows. The predicted leader trajectory is also drawn (purple), and we can see that the robot follows it while still planning its own path around the obstacles. As described in Section~\ref{sec:leader-following}, the robot only plans toward the current leader's relative following position, letting waverider handle the rest, and does not explicitly backtrack the leader's path. We also repeat this experiment while having someone else block the robot's path. In all but one experiment, the robot dodged the person and continued tracking the leader. In the one case, it temporarily locked onto them as a leader but quickly noticed the mistake from the AoA beacon readings (see Section~\ref{sec:leader-tracking}) and corrected itself. In such cases it only takes a few seconds until the robot is once again headed towards the correct leader.

\begin{figure}[!t]
    \centering
    \includegraphics[width=1.0\linewidth]{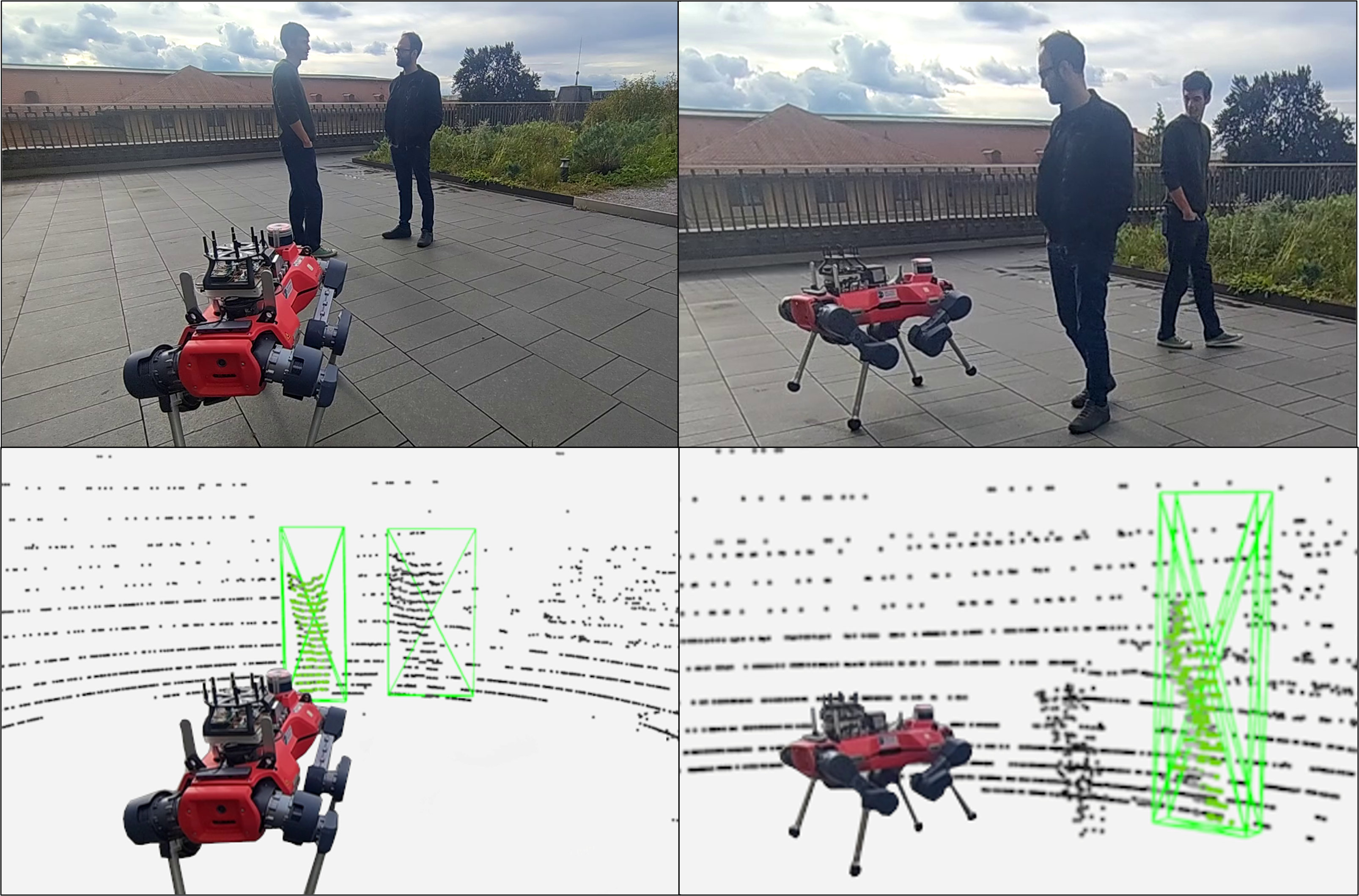}
    \caption{In a situation with two people present, the pipeline correctly segments out both (green bounding boxes), and identifies its leader from the two. The leader point cloud is additionally highlighted in green. When the leader starts to move, as shown on the right, the robot follows along, diverging to the left to avoid collision with the other human.}
    \label{fig:results}
    \vspace{-0.5cm}
\end{figure}

\subsubsection{Leader Selection}
We evaluate the leader selection and tracking process when multiple people stand beside each other, as shown in Figure~\ref{fig:results}. Our AoA sensor allows us to easily tackle this common-place scenario. In the experiment, we initialized the robot with multiple people in front of it. Throughout multiple runs, it was always able to correctly predict the leader and, upon leader motion, start following them. Even in cases when the leader's path went around the other person, our tracking system maintained the correct target throughout the temporary occlusion.

\subsubsection{Dynamic Indoor Scenes}
Finally, we also show robustness in complex scenarios. We perform leader following in indoor scenarios, which are challenging due to the increased number of obstacles and the degradation in signal quality from the AoA beacon. Simultaneously, we also have people interfere with the robot, coming between the leader and the robot or forcing it to dodge them to continue its path. In some instances, the leader tracking temporarily switches to one of the passing people. However, the robot can recover and continue tracking the correct leader, showcasing the resilience of our proposed method. This shows we are able to address and overcome issues and challenges described by existing state-of-the-art research~\cite{Fang2010ABeacon, Afghani2013FollowBeacons, Tripathi2021HumanSensor, Yang2019ASystem, Pena2019FollowOptimization}.

\section{CONCLUSIONS}

In this paper we have presented a end-to-end pipeline for a quadrupedial robot, enabling it to follow a human leader through dynamic environments. The solution is minimally invasive for the human operator, as it only requires them to carry a lightweight, pocket-sized transmitter beacon. Robust multi-sensor fusion allows us to navigate both in open spaces and through moving crowds, without loosing track of the operator. Our system enables robots to move into more commonplace settings, as it reduces the burden of operator training. This is particularly interesting for personal assistance robots, such as those that could extend the mobility of people with disabilities or for applications where robots assist in the inspection of industrial sites.

\addtolength{\textheight}{0cm}   




\bibliographystyle{IEEEtran}
\bibliography{root.bib}

\end{document}